\documentclass{article} 
\usepackage{iclr2026_conference,times}
\usepackage{subcaption}

\usepackage{amsmath,amsfonts,bm}

\def\eqref#1{equation~\ref{#1}}

\def\1{\bm{1}}

\DeclareMathAlphabet{\mathsfit}{\encodingdefault}{\sfdefault}{m}{sl}
\SetMathAlphabet{\mathsfit}{bold}{\encodingdefault}{\sfdefault}{bx}{n}

\usepackage{multirow}
\usepackage{hyperref}

\usepackage{url}
\usepackage{graphicx} 
\usepackage{amsmath}
\usepackage{xcolor} 
\usepackage{todonotes}
\usepackage{algorithm}
\usepackage{algpseudocode}

\newcommand{\ourmodelacronym}{ETT~}
\newcommand{\longsequnce}{s}
\newcommand{\token}{t}
\newcommand{\llminput}{\mathcal{C}}

\newcommand{\overlap}{\delta}

\newcommand{\chunklen}{\ell}

\newcommand{\batch}{\mathcal{B}}
\newcommand{\batchi}{i}

\newcommand{\T}{E}
\usepackage{arydshln}
\newcommand{\question}{I}

\newcommand{\parameter}{\theta}

\title{ETT: Expanding the Long Context Understanding Capability of LLMs at Test-Time}

\author{Kiarash Zahirnia, Zahra Golpayegani, Walid Ahmed\thanks{ walid.ahmed1@huawei.com}, \& Yang Liu  \\
Ascend Team, Toronto Research Center, Huawei Technologies
}

\iclrfinalcopy 
\begin{document}

\maketitle

\begin{abstract}
Transformer-based Language Models' computation and memory overhead increase quadratically as a function of sequence length. The quadratic cost poses challenges when employing LLMs for processing long sequences. In this work we introduce \ourmodelacronym~(Extend at Test-Time), a method that extends the context length of short-context Transformer-based LLMs  with constant memory requirement and linear computation overhead.
\ourmodelacronym~enable the extension of  context length at test-time by efficient fine-tuning the model parameters on the input context, chunked into overlapping small subsequences.

We evaluate \ourmodelacronym~on LongBench by extending the context length of GPT-Large and Phi-2 up to 32 times, increasing from 1k to 32k tokens. This results in up to a 30\% improvement in the model's accuracy.
We further investigate how contextual information can be effectively stored within LLM parameters. 
Through a detailed ablation study, we examine which Transformer modules are most beneficial to fine-tune at test-time.
Interestingly, we find that fine-tuning the second layer of the FFNs is more effective than full fine-tuning, leading to a further improvement in the models' accuracy.
\end{abstract}

\section{Introduction}

Transformers have achieved state-of-the-art performance across a wide range of tasks \cite{vaswani2023attentionneed}. Nonetheless, their scalability to long sequences is fundamentally constrained by the computational and memory demands of the attention mechanism. In particular, standard attention incurs quadratic complexity $\mathcal{O}(N^2)$ in both computation and memory with respect to sequence length $N$, while even optimized variants such as flash attention exhibit linear memory growth \cite{dao2022flashattention}. Moreover, during inference, the key–value (KV) cache expands proportionally with the sequence length, introducing additional memory bottlenecks that further impede efficient long-context processing.

In this work, we investigate Test-Time Training (TTT) \cite{krause2017dynamicevaluationneuralsequence} to extend the model's context length at test (inference) time with constant memory requirements and linear computational complexity. TTT updates the model parameters using a loss derived by unlabeled test data, and resets the model parameters to their original value after completing the inference for each test data. 
We introduce \ourmodelacronym~(Extend at Test-Time), which extend the context length at test-time by fine-tuning the model's parameters on the input context, chunked into overlapping subsequences.

\emph{From a memory perspective}, \ourmodelacronym~leverages the model's parameters and their ability to memorize the data as persistent memory during inference, resetting them at the end of the process. \ourmodelacronym~reduces the computational overhead of transformer based LLMs from quadratic to linear and maintains a constant memory footprint regardless of input length since the model input is limited to fixed chunk size.

Our primarily empirical experiment investigates extending the short-context window of small language models (Phi-2 \cite{javaheripi2023phi} and GPT-Large \cite{radford2019language}) by up to 32× at test-time through full fine-tuning. This approach result in a noticeable improvement in LongBench \cite{bai2024longbench} scores. 




While \ourmodelacronym~has a constant memory requirement, (full) test-time training incurs a 3× model-size overhead, primarily due to the need to store optimizer states and gradients. This raises an important question: \emph{Can we efficiently  “memorize” the input context at test-time?}
To explore this, we conduct empirical studies focused on two key aspects: (1) which model modules, such as self-attention or feed-forward networks, are most effective to fine-tune, and (2) whether fine-tuning shallower versus deeper layers leads to better performance on long-context understanding tasks.

\noindent We conduct an empirical ablation study on fine-tuning FFNs (also known as key-value memories \cite{geva2021transformer}), keys (the 1st layer in the FFNs), values (the 2nd layer in the FFNs), and attention layers and compare them with full fine-tuning. We compare those methods
in various long-context understanding tasks and generally observe the superiority of fine-tuning keys over other modules, including full fine-tuning. In fact, we observe that TTT on only key parameters improves the model accuracy while substantially reduces the learnable parameters.

\noindent We also empirically evaluate the effectiveness of shallower key layers in \ourmodelacronym~performance and observe that shallow layers contribute minimally
to the overall performance. In fact, our result shows that we can remove a fraction of the shallower layers from Test-Time Training parameters with minimal degradation in downstream Long Context Understanding benchmarks. This finding allows us to reduce the overhead of applying \ourmodelacronym~ by freezing the shallow layers and limiting the back-propagation through a the remaining deep layers.

To summarize, our contributions are the following:
\begin{itemize}

\item We propose \ourmodelacronym, an architecture-agnostic method that extends the context length of short context pretrained language models at test-time with constant memory and linear computational overhead.

\item Through ablation studies, we find that fine-tuning only the first layer of FFN modules (key layer) is more effective than full model tuning, reducing the overhead while improving the performance. Furthermore, we show that training only the top (deeper) layers  of the model preserves performance while reducing compute and memory costs.
    
\end{itemize}

The rest of this paper is organized as follows: Section \ref{sec:related-work} provides some context about the related work. Section \ref{sec:method} describes \ourmodelacronym~in detail. In Section \ref{sec:experiments}, we highlight the experiments, and finally, we conclude our findings in Section \ref{sec:conclusion}.
\section{Related Work}
\label{sec:related-work}

Several efforts have been made to overcome the quadratic memory bottleneck in Transformers. Sparse attention mechanisms selectively limit which tokens should participate in self-attention, reducing the complexity from quadratic to linear or sub-quadratic levels depending on the sparsity pattern \cite{child2019generatinglongsequencessparse, beltagy2020longformerlongdocumenttransformer}. While sparse attention-based methods can successfully increase the context length by reducing the complexity, they rely on predefined attention patterns. Kernel-based methods \cite{katharopoulos2020transformersrnnsfastautoregressive} address the challenge of quadratic complexity by approximating the Softmax function in self-attention with a kernel function, enabling attention computation with linear complexity. However, despite their efficiency, kernel-based methods fall short of Softmax attention both in terms of accuracy and training stability \cite{qin2022devillineartransformer}. Alternative architectures to Transformers, including recurrent architectures such as State Space Models (SSMs) \cite{gu2024mambalineartimesequencemodeling} and State Space Duality (SSD) \cite{dao2024transformersssmsgeneralizedmodels}, have been proposed to address the quadratic costs at the architectural level and enable scalable evaluation over long-contexts with linear complexity. However, these models often suffer from limited expressiveness \cite{chen2025computationallimitsstatespacemodels} due to their fixed-size hidden states, which constrains their ability to capture complex dependencies and ultimately leads to lower accuracy compared to Transformers in long-context evaluation.

TTT has a long-standing history in the field of machine learning \cite{Hinton1987UsingFW, bottou1992local, schmidhuber1992learning}. 
Recently, TTT has been revisited by researchers to be applied to language modeling \cite{NIPS2016_9f44e956, hübotter2025efficientlylearningtesttimeactive, sun2025learninglearntesttime, hardt2024testtimetrainingnearestneighbors, mahdi2025continuous}.
 The basic approach is to directly fine-tune a language model on the test
sequence to learn the local probability distribution.
Dynamic Evaluation \cite{pmlr-v80-krause18a} fine-tunes the model parameters during training with a next-word prediction objective function and substantially improves the model's perplexity. However, it requires over three times the computational cost compared to standard inference. Authors in \cite{clark2022metalearningfastweightlanguage} improve the efficiency of Dynamic Evaluation by adding a linear layer, called Fast Weight Layer (FWL), on top of the existing transformer models and only fine-tuning the FWL at test-time.
While Dynamic Evaluation and FWL has shown perplexity improvements, their performance on downstream tasks remains unexplored.
In this work, we explore the effectiveness of TTT for improving the long-context understanding capabilities of large language models (LLMs) with constant memory requirement.

In a concurrent work, LIFT \cite{mao2025lift} proposed memorizing the context in a specialized Gated Memory and utilizing auxiliary tasks, handcrafted for each downstream task, to fine-tune the model at test-time and improve LLMs' long-context performance. In contrast, \ourmodelacronym~fine-tunes a subset of the model parameters using a next-word prediction objective function and empirically demonstrates that TTT can effectively and efficiently improve the LLM's long-context understanding capability without the need for external memory or auxiliary task design.

\section{Method}
\label{sec:method}

At test (inference) time, given a prompt consisting of an instruction $\question$~and a long context $\llminput$, \ourmodelacronym~fine-tunes the pretrained model with parameter $\parameter_0$ on the long context $\llminput$ and implicitly memorizes the sequence in the model parameters. To address the quadratic computation overhead and memory footprint of transformer based models, \ourmodelacronym~chunks long context $\llminput = (\token_0, \token_1,\dots,\token_L)$ into overlapping subsequences $\{\longsequnce_0 = \token_{0:\chunklen}, \longsequnce_1=\token_{\chunklen-\overlap:2\chunklen}, ... \}$, with fixed length of $\chunklen+\overlap$ tokens. The subsequences are randomly grouped into batches, with batch $i$ (zero-indexed) denoted as $\batch_\batchi$, and fine-tuned using a next-word prediction objective function to edit the model's implicit knowledge.


The pretrained model parameters are used to compute the log probability of the first batch $\sum_{{\longsequnce_\batchi \in \batch_0}} \log p(\longsequnce_\batchi|\theta_0)$. This probability is then employed to calculate the cross-entropy loss 
$\mathcal{L}_{\text{NWP}}(\batch_0)$ and the corresponding gradient $\nabla \mathcal{L}_{\text{NWP}}(\batch_0)$. The gradient $\nabla \mathcal{L}_{\text{NWP}}(\batch_0)$ is subsequently used to update the model, resulting in the adapted parameters 
$\theta_1$. This process is repeated for the second batch, where the probability 
$p(\batch_1|\theta_1)$ is evaluated, and the procedure is carried out iteratively for the remaining batches (See Algorithm \ref{al:1}).

\begin{algorithm}
\caption{\ourmodelacronym~Algorithm}
\begin{algorithmic}[1]
\State \textbf{Input:} Pretrained model $\mathcal{M}$ with parameters $\parameter_0$, Context sequence $\llminput$, Instruction $\question$, chunk length $\llminput$, overlap length  $\overlap$, number of TTT epochs \T.
\State \textbf{Output:} Generated answer $A$.
\vspace{1em}
\State Chunk, $\llminput$ into overlapping subsequences: 
$\{\longsequnce_0 = \token_{0:\chunklen}, \longsequnce_1=\token_{\chunklen-\overlap:2\chunklen}, ... \}$
\For{each epoch e $\in [1, \T]$}
\State Randomly group subsequences into batches, batch $i$ denoted as $\batch_\batchi$
\For{each batch $\batch_\batchi$}
    \State Fine-tune model parameters:
        \[
        \mathcal{M}_{\boldsymbol{\theta}_e} \leftarrow \text{Update}(\mathcal{M}_{\boldsymbol{\theta}_{e-1}}, \mathcal{L}_{\text{NWP}}(\mathcal{B}_i)),
        \]
        where $\mathcal{L}_{\text{NWP}}$ is the next-word prediction loss.
\EndFor
\EndFor
\State Generate answer: $A \sim p_{{\theta}_\T}(.|\question)$.
\State Reset model parameters.
\State \Return $A$.
\end{algorithmic}
\label{al:1}
\end{algorithm}

\section{Experiments}
\label{sec:experiments}
We evaluate \ourmodelacronym~on GPT-Large and Phi-2.  We evaluate \ourmodelacronym on LongBench \cite{bai2024longbench}, which comprises 21 real-world and synthetic long-context tasks.

We begin by examining the improvements in long-context capabilities of the studied models with \ourmodelacronym~and full fine-tuning at test-time. Next, we investigate whether the test-time training overhead can be reduced  by selective fine-tuning. Specifically, we demonstrate that: 1) Fine-tuning only the up-projection layers in the feed-forward networks (also known as key \cite{geva2021transformer}) can further improve accuracy compare to full fine-tuning while reducing the number of trainable parameters by approximately 70\%.
2) We find that restricting fine-tuning to only the deeper layers allows us to reduce the number of trainable parameters at test-time to just 15\% of the model's parameters, with little to no loss in performance.

\paragraph{Experimental details.}
In all of the experiments, we chunk the long-context input into subsequences of 512 tokens with an overlap of 32 tokens between the adjacent chunks. For each input, we fine-tune the model for 10 epochs and restore the original model parameters after running inference. We adopt the Adam optimizer with a learning
rate of $5e^{-4}$ and weight decay of $0.5$.

\subsection{\ourmodelacronym~Enhances Long-Context Understanding Across Standard Long-Context Tasks}
\begin{figure}[h!]
    \centering
    \begin{subfigure}[b]{0.325\textwidth}
        \centering
        \includegraphics[width=\textwidth]{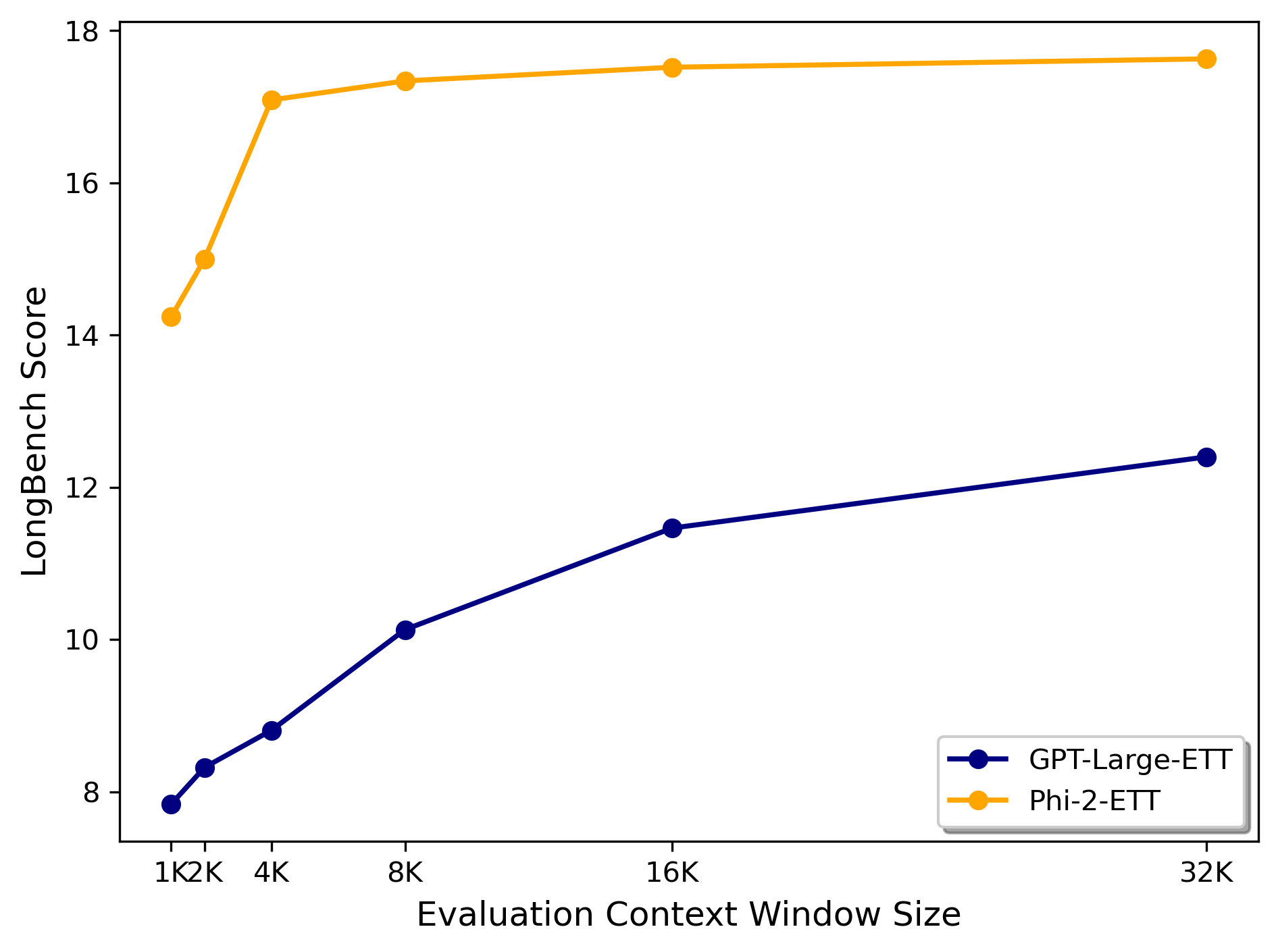}
        \caption{Average LongBench Score}
        \label{fig:sub1}
    \end{subfigure}
    \hfill
    \begin{subfigure}[b]{0.325\textwidth}
        \centering
        \includegraphics[width=\textwidth]{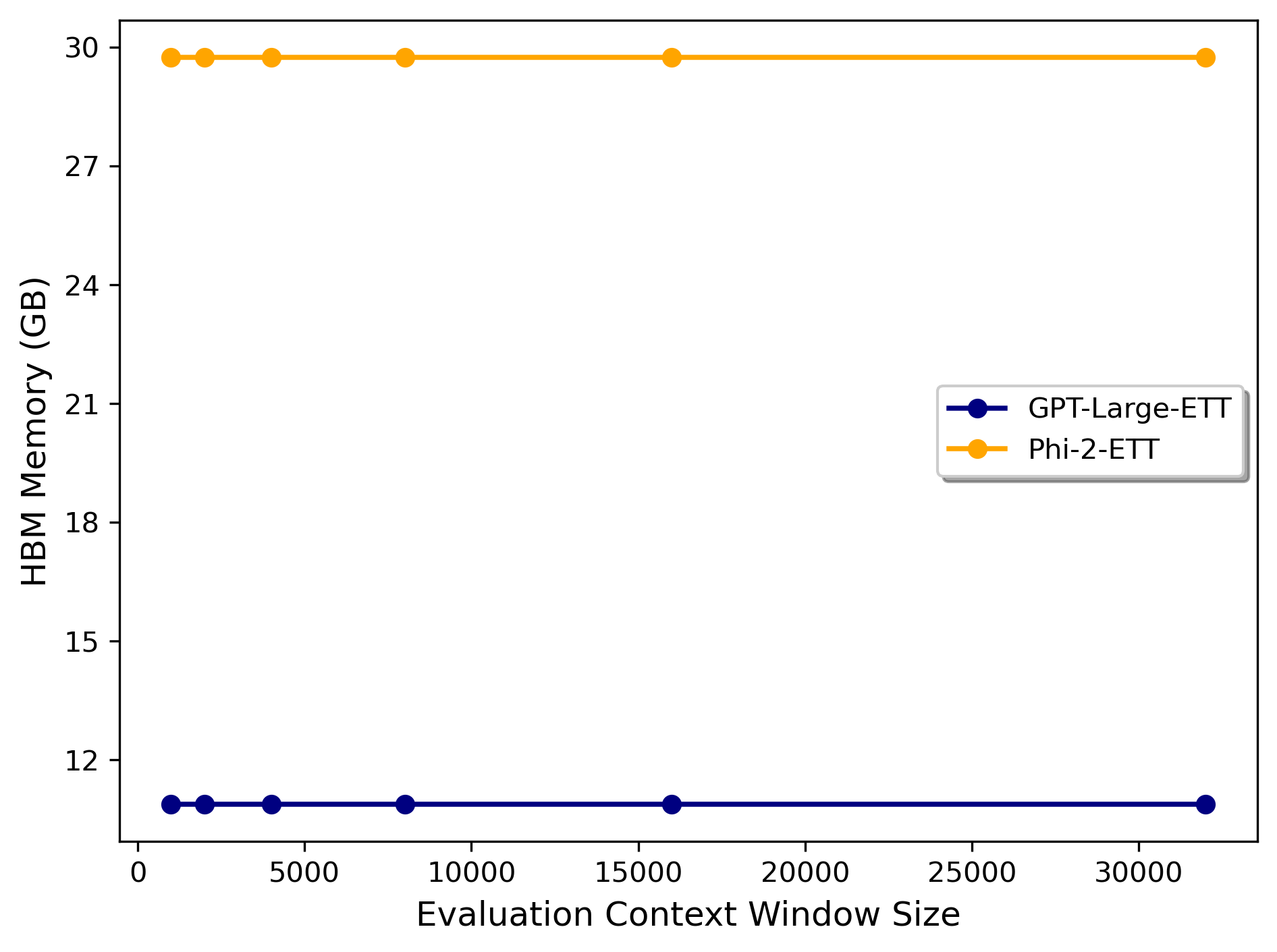}
        \caption{HBM Memory Usage}
        \label{fig:sub2}
    \end{subfigure}
    \hfill
    \begin{subfigure}[b]{0.325\textwidth}
        \centering
        \includegraphics[width=\textwidth]{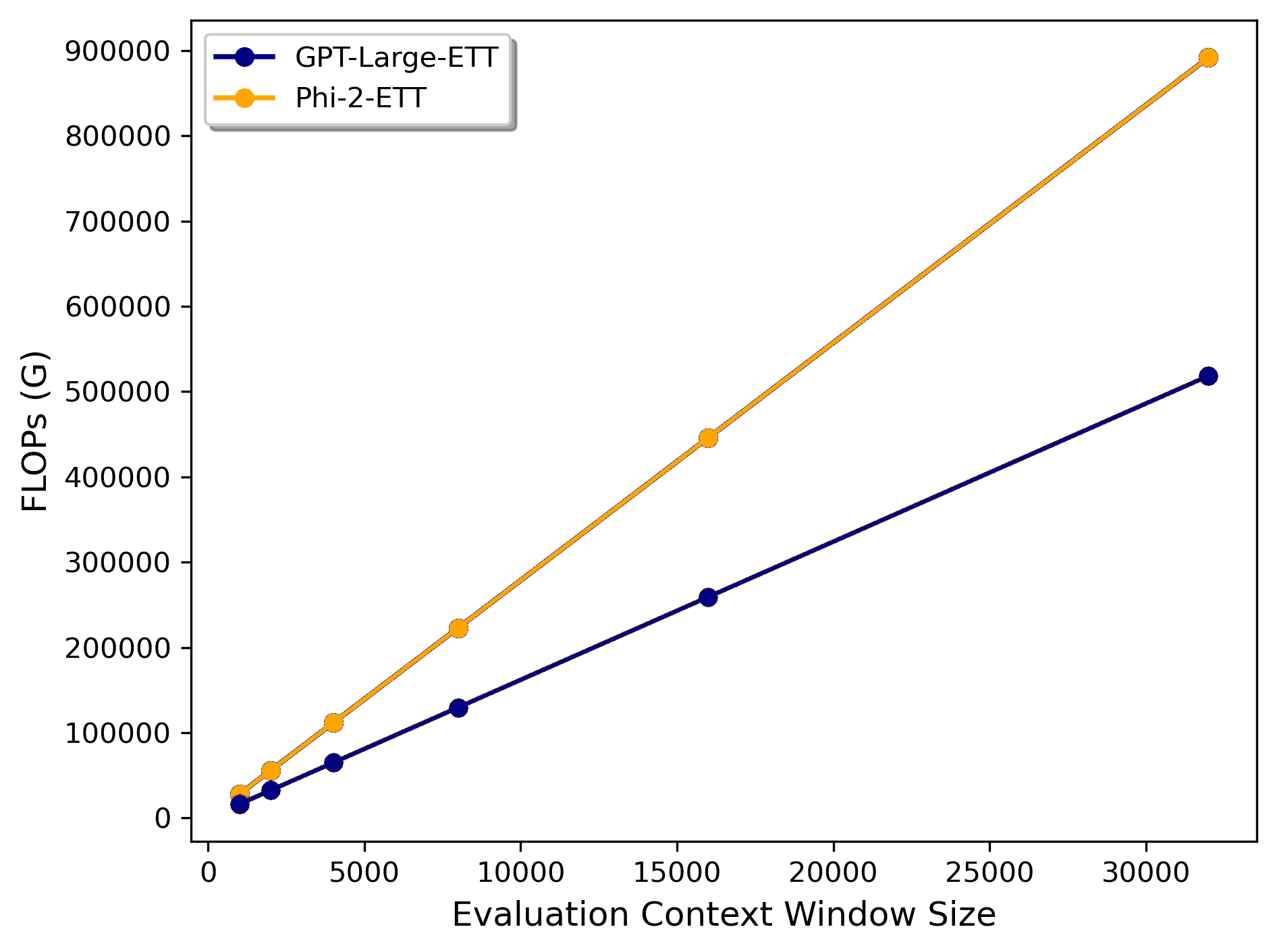}
        \caption{FLOPs}
        \label{fig:sub3}
    \end{subfigure}
    
    \caption{Average LongBench score (\%), HBM memory consumption (GB), 
    and FLOPs (G) under different truncation sizes. 
    \ourmodelacronym~extends the context window of Phi-2 and GPT-Large by up to 16× and 32×, respectively. 
    Performance improves with longer context lengths while maintaining constant memory usage and only linear growth in computation.}
    \label{fig:scaleLaw}
\end{figure}

Figure~\ref{fig:scaleLaw} illustrates how \ourmodelacronym~affects the long-context understanding abilities of Phi-2 and GPT-Large as a function of context length. In all experiments, the context $\llminput$ is truncated in the middle, following the setup of \citet{bai2024longbench}. We apply full fine-tuning at test time and report the average LongBench score across all 21 tasks. The results show a consistent improvement in performance across all tasks as the context length increases.

All experiments are conducted on a single NVIDIA V100 GPU with 32 GB of HBM2 memory. The memory footprint remains constant across different context window sizes. We estimate the training FLOPs for full fine-tuning following \citet{kaplan2020scaling}. For HBM usage, we report the observed memory consumption during full fine-tuning and training on a single chunk of 512 + 32 tokens using BF16 mixed precision and the Adam optimizer with gradient accumulation.

\subsection{Selective Fine-Tuning at Test-Time Outperforms Full Fine-Tuning}
In this work, we conduct an empirical ablation study to evaluate the effectiveness of selectively fine-tuning different modules in enhancing long-context understanding at test-time. Specifically, we fine-tune individual modules of the model: the keys (i.e., the first linear layer in the FFN, denoted as $\text{FFN}_{\text{Up}}$), the values (i.e., the second linear layer in the FFN, denoted as $\text{FFN}_{\text{Up}}$), and the attention parameters (i.e., the key, query, and value projections: K, Q, V). We compare these strategies based on their impact on \ourmodelacronym's performance.

\emph{This experiment aims to provide insights into the effectiveness of fine-tuning different modules at test-time.} 

As shown in Table~\ref{tbl:targetedFineTunning}, fine-tuning $\text{FFN}_{\text{Up}}$ consistently outperforms other strategies across various settings. In particular, fine-tuning $\text{FFN}_{\text{Up}}$ instead of applying full fine-tuning improves the LongBench score from 11.30 to 12.57 for GPT-Large, and from 16.75 to 18.3 for Phi-2 while reducing the number of trainable parameters—and consequently the memory footprint—by 70\%. This observation aligns with previous studies, which have shown that updating the keys within FFNs leads to performance improvements compared to updating the values when tuning LLMs for knowledge editing task \cite{qiu2024empirical}.

\begin{table}[ht]
\centering
\resizebox{\textwidth}{!}{
\begin{tabular}{l c c c c c}
\hline
\textbf{ETT Target} & \multicolumn{2}{c}{\textbf{GPT-Large \cite{radford2019language}}} & \multicolumn{2}{c}{\textbf{Phi-2 \cite{javaheripi2023phi}}} \\
\cline{2-3} \cline{4-5}
 & \textbf{Trainable} & \textbf{LongBench Score} & \textbf{Trainable} & \textbf{LongBench Score} \\
\hline
Full Fine-Tuning & 100.0 \% & 11.30 & 100.0 \% & 17.33 \\
\hline
FFN & 60.99 \% & 11.81 & 60.37 \% & 17.21 \\
\hline
$\text{FFN}_{\text{Up}}$ & 30.48 \% & \textbf{12.57} & 30.19 \% & \textbf{18.33}\\
\hline
$\text{FFN}_{\text{Down}}$ & 30.50 \% & 11.15 & 30.18 \% & 16.75 \\
\hline
$\text{Attention}_{\text{QKV}}$ & 30.48 \% & 11.11 & 30.19 \% & 18.31 \\
\hline
Baseline & 0 \% & 9.58 & 0 \% & 15.04 \\
\hline
\end{tabular}
}
\caption{ETT Target and corresponding LongBench scores for Experiment GPT-Large and Phi-2.}
\label{tbl:targetedFineTunning}
\end{table}
\subsection{Shallower Key Layers Are Less Effective Than The Deeper Ones
}
We also empirically investigate the effectiveness of fine-tuning shallower $\text{FFN}_{\text{Up}}$ layers at test-time. If we freeze a block of shallow layers and observe no impact on \ourmodelacronym's performance, it suggests that those layers are not essential for \ourmodelacronym. To identify the optimal block of shallow layers to freeze, we incrementally freeze blocks of shallow layers and evaluate \ourmodelacronym's performance at each step. This bottom-up strategy reduces the number of trainable parameters and computational cost as backpropagation is not required for the contiguous block of shallow, frozen layers.

\begin{figure}[h!]
    \centering
    \includegraphics[width=0.5\textwidth]{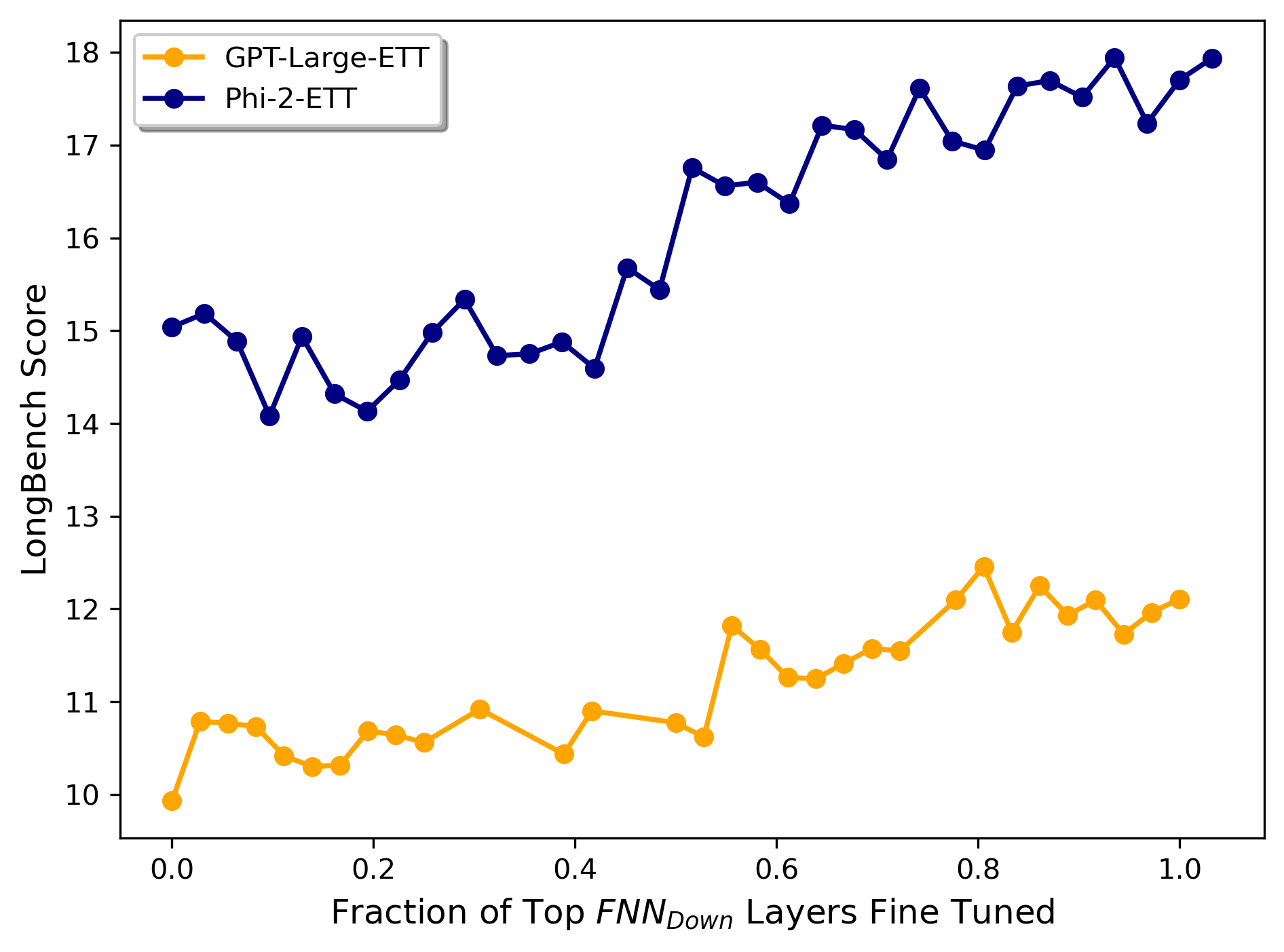}  
    \caption{\ourmodelacronym's LongBench score as a function of the fraction of deep $\text{FFN}_{\text{Up}}$ layers fine-tuned. We can store the long input in the parameters of the top 80\% of $\text{FFN}_{\text{Up}}$ layers without significant performance degradation.}
    \label{fig:prunning}
\end{figure}

Figure~\ref{fig:prunning} shows \ourmodelacronym's average LongBench score as the fraction of shallow key ($\text{FFN}_{\text{Up}}$) layers frozen. We observe that fine-tuning only the top 80\% of $\text{FFN}_{\text{Up}}$ layers achieves similar performance as fine-tuning all layers. Importantly, there is a sharp performance degradation when freezing more than 40\% of the shallow layers, indicating a transition point beyond which key contextual information is no longer preserved.

The LongBench scores for GPT-Large and Phi-2, with and without the parameter-efficient version of \ourmodelacronym, are reported in Tables~\ref{tbl.longbench_part1} and ~\ref{tbl.longbench_part2}. In all the experiments, we fine-tuned the top 80\% of the $\text{FFN}_{\text{Up}}$ layers.

\begin{table}[h!]
\centering
\resizebox{\textwidth}{!}{%
\begin{tabular}{c c c c c c c c c c c c c}
& \multicolumn{4}{c}{\textbf{Single-Doc QA}} 
& \multicolumn{4}{c}{\textbf{Multi-Doc QA}} 
& \multicolumn{4}{c}{\textbf{Summarization}} \\ \cline{2-13}

& \rotatebox{90}{\textbf{MultiFieldQA-zh}} 
& \rotatebox{90}{\textbf{MultiFieldQA-en}} 
& \rotatebox{90}{\textbf{NarrativeQA}} 
& \rotatebox{90}{\textbf{Qasper}} 
& \rotatebox{90}{\textbf{2WikiMultihopQA}} 
& \rotatebox{90}{\textbf{HotpotQA}} 
& \rotatebox{90}{\textbf{MuSiQue}} 
& \rotatebox{90}{\textbf{DuReader}} 
& \rotatebox{90}{\textbf{MultiNews}} 
& \rotatebox{90}{\textbf{GovReport}} 
& \rotatebox{90}{\textbf{QMSum}} 
& \rotatebox{90}{\textbf{VCSUM}} \\ \hline

\textbf{GPT-Large}& \textbf{6.7}& \textbf{13.36 }& 2.2& 5.29   & \textbf{9.3}& 5.55   & 3.1& 14.19  & 21.62& 19.01& 13.41& 8.67 \\ 
\textbf{GPT-Large-\ourmodelacronym}& 5.62  & 12.07 & \textbf{2.7}& \textbf{7.22}   & 8.52   & \textbf{6.47}   & \textbf{4.54}   & \textbf{17.36}  & \textbf{24.81}& \textbf{25.99}& \textbf{13.96}& \textbf{9.53} \\
\hdashline
\textbf{Phi-2}&\textbf{19.42} & 34.9 & \textbf{13.12} & 6.94 & 9.27 & 12.3 & 9.14 & 6.54 & \textbf{32.52} & 26.32 & {18.09} & \textbf{11.06} \\
\textbf{Phi-2-\ourmodelacronym}&18.37 & \textbf{37.76} & 9.46 & \textbf{9.64} & \textbf{13.3} & \textbf{18.26} & \textbf{9.26} & \textbf{20.2} & 25.5 & \textbf{34.01} & \textbf{19.3} & 10.55 \\
\hline
\end{tabular}
}
\caption{LongBench score comparison between GPT-Large and Phi-2, with and without \ourmodelacronym~(selectively fine-tuned) on {Single-Doc QA}, {Multi-Doc QA}, and {Summarization} tasks.}
\label{tbl.longbench_part1}
\end{table}

\begin{table}[h!]
\centering
\resizebox{\textwidth}{!}{%
\begin{tabular}{c c c c c c c c c c c}
& \multicolumn{4}{c}{\textbf{Few-shot Learning}} 
& \multicolumn{3}{c}{\textbf{Synthetic Tasks}} 
& \multicolumn{2}{c}{\textbf{Code Completion}} 
& \textbf{} \\ \cline{2-11}

& \rotatebox{90}{\textbf{TriviaQA}} 
& \rotatebox{90}{\textbf{SAMSum}} 
& \rotatebox{90}{\textbf{TREC}} 
& \rotatebox{90}{\textbf{LSHT}} 
& \rotatebox{90}{\textbf{PassageRetrieval-en}} 
& \rotatebox{90}{\textbf{PassageRetrieval-zh}} 
& \rotatebox{90}{\textbf{PassageCount}} 
& \rotatebox{90}{\textbf{LCC}} 
& \rotatebox{90}{\textbf{RepoBench-P}} 
& \rotatebox{90}{\textbf{Avg}} \\ \hline

\textbf{GPT-Large}& \textbf{10.65}& 22.12& 22.83  & 0 & \textbf{3.47}  & \textbf{2.5}   & 0.87  & 9.12  & 13.06 & 9.85\\ 
\textbf{GPT-Large-\ourmodelacronym}& 7.56 & \textbf{24.42}& \textbf{27.98}  & \textbf{4.5}   & 3.33  & 1.83  & \textbf{1.9}   & \textbf{25}   & \textbf{22.52} & \textbf{12.27}\\
\hdashline
\textbf{Phi-2}&2.38 & 3.03 & 28.57 & \textbf{23.15} & {14.29} & {4.76}  & \textbf{1.59} & \textbf{18.05} & {20.52} & 15.04\\
\textbf{Phi-2-\ourmodelacronym}& \textbf{2.38} & \textbf{15.38} & \textbf{42.86} & 22.53 & \textbf{14.29} & \textbf{19.05} & {0} & {16.1} & \textbf{22.29} & \textbf{18.34}\\
\hline
\end{tabular}
}
\caption{LongBench score comparison between GPT-Large and Phi-2, with and without \ourmodelacronym~(selectively fine-tuned) on {Few-shot Learning}, {Synthetic}, and {Code Completion} tasks.}
\label{tbl.longbench_part2}
\end{table}
\newpage
\subsection{\ourmodelacronym Enables Phi-2 to Compete with 8B LLMs Fine-Tuned on Long Contexts, Using Constant Memory and Linear Computation}
We further compare ETT against several popular baselines on the long-context benchmarks, including fixed-length models fine-tuned on long-context data (Vicuna1.5-7B-16k\footnote{\url{https://huggingface.co/lmsys/vicuna-7b-v1.5-16k}}, LongChat1.5-7B-32k\footnote{\url{https://huggingface.co/lmsys/longchat-7b-v1.5-32k}}, together/llama-2-7b-32k\footnote{\url{https://huggingface.co/togethercomputer/LLaMA-2-7B-32K}}, Llama-3-8B-Instruct-Gradient-1M\footnote{\url{https://huggingface.co/gradientai/Llama-3-8B-Instruct-Gradient-1048k}}), as well as context-extension methods such as SelfExtend \cite{jin2024llm} and LIFT \cite{mao2025lift}. We benchmark the models on five  long-context tasks from LongBench, using the same tasks reported in LIFT.

Across almost all tasks, when applied to Phi-2, ETT consistently outperforms context-extension methods. Specifically, on PassageRetrievalEN and Musique, Phi-2-ETT achieves 14.29 and 9.26, respectively, substantially exceeding the performance of Phi-2-LIFT (8.17, 3.96) and Phi-2-SelfExtend (2.38, 3.89).

 ETT also allows Phi-2 to achieve competitive results relative to 8B-parameter models despite having significantly fewer parameters. Notably, \ourmodelacronym improves Phi-2’s performance on GovReport from 26.32 to 34.01, surpassing all studied baselines.

Overall, these results confirm that ETT substantially enhances the long-context capability of Phi-2, outperforming SelfExtend and LIFT on several key benchmarks, and achieving performance competitive with much larger LLMs.
\begin{table}[h!]
\centering
\resizebox{\textwidth}{!}{%
\begin{tabular}{lcccccc}
\hline
\textbf{Methods} & \textbf{Musique} & \textbf{Narrativeqa} & \textbf{Qmsum} & \textbf{GovReport} & \textbf{PassageRetrievalEN} \\
\hline
{{Phi-2-{ETT}} } (ours*)       & \textbf{9.26} & 9.46 & \textbf{19.30} & \textbf{34.01} & \textbf{14.29} \\
{Phi-2-LIFT}        & 3.96 & 11.78 & 15.32 & 29.39 & 8.17 \\
{Phi-2-Se}     &  3.89   & \textbf{12.04} & 14.58 & 27.90 & 2.38 \\
\hline

LLaMa3-8B-32k-LIFT   & 10.99 & \textbf{25.84} & \textbf{22.96} & \textbf{31.26} & \textbf{41.67} \\

LLaMa3-8B-32k-Se   & 3.89 & 12.04 & 14.58 & 27.90 & 2.83\\
\hdashline
Llama-3-8B-Instruct-Gradient-1M    & \textbf{13.89} & 12.04 & 14.58 & 27.90 & 2.83 \\
together/llama-2-7b-32k  & 6.19 & 15.65 & 17.18 & 29.28 & 23.0 &  \\
Vicuna1.5-7B-16k & 9.8 & 19.4 & 22.8 & 27.9& 4.5 \\
LongChat1.5-7B-32k  & 9.7 & 16.6 & 22.7 & 30.08 &  30.50\\
\hline
\end{tabular}}
\caption{Performance comparison of different LLMs on LongBench. The number (e.g., ‘25k’) denotes the maximum input length. The postfixes Se, LIFT, and ETT indicate that SelfExtend \cite{jin2024llm}, LIFT \cite{mao2025lift}, and ETT (our method), respectively, are applied to the corresponding model. LongChat1.5-7B-32k , together/llama-2-7b-32k and Vicuna1.5-7B-16k are fixed-length models fine-tuned on long contexts \cite{jin2024llm}. The best performance is highlighted in bold. \ourmodelacronym enhances the long-context understanding of Phi-2 (2.7B parameters) to a level competitive with models up to 8B parameters, while maintaining constant memory requirements and linear computation with respect to context length. Notably, \ourmodelacronym improves Phi-2’s performance on GovReport from 26.32 to 34.01, surpassing all studied baselines.}
\end{table}

\section{Conclusion and Future Work}
\label{sec:conclusion}
In this work, we introduce \ourmodelacronym, an architecture-agnostic, lightweight and efficient approach for extending the context length of pretrained language models at inference time with constant memory and linear computation overhead. Our method enables transformer based language models, such as GPT-Large and Phi-2, originally trained with short context windows to process significantly longer inputs. \ourmodelacronym~demonstrates consistent improvements in long-context understanding across multiple tasks from LongBench. We also investigated the effectiveness of different transformer modules and shallow-layer in test-time training. Specifically, we demonstrated that: 1) Fine-tuning only the up-projection layers in the feed-forward networks improves ETT accuracy compared to full fine-tuning while reducing the number of trainable parameters by approximately 70\%. 2) We showed that restricting fine-tuning to only the deeper layers allows us to reduce the number of trainable parameters at test-time to just 15\% of the model’s parameters, with little to no loss in performance.
Our results highlight the effectiveness of \ourmodelacronym, offering a practical solution for scaling LLMs to longer sequences.
\newpage
\bibliography{iclr2026_conference}
\bibliographystyle{iclr2026_conference}

\end{document}